# Exploring Equality:
# An Investigation into Custom Loss Functions for Fairness Definitions

**Gordon Lee***, Simeon Sayer

gordonlee300@gmail.com, ssayer@fas.harvard.edu

*****Primary Author

---

**I. Abstract**

    This paper explores the complex tradeoffs between various fairness metrics—such as equalized odds, disparate impact, and equal opportunity—and predictive accuracy within COMPAS by building neural networks trained with custom loss functions optimized to specific fairness criteria. This paper creates the first fairness-driven implementation of the novel Group Accuracy Parity (GAP) framework, as theoretically proposed by Gupta et al. (2024) [1], and applies it to COMPAS. To operationalize and accurately compare the fairness of COMPAS models optimized to differing fairness ideals, this paper develops and proposes a combinatory analytical procedure that incorporates Pareto front and multivariate analysis, leveraging data visualizations such as violin graphs. This paper concludes that GAP achieves an enhanced equilibrium between fairness and accuracy compared to COMPAS's current nationwide implementation and alternative implementations of COMPAS optimized to more traditional fairness definitions [2][3][4][5][6][7].
    While this paper's algorithmic improvements of COMPAS significantly augment its fairness, external biases undermine the fairness of its implementation. Practices such as predictive policing [8] and issues such as the lack of transparency regarding COMPAS's internal workings [4][9][10][11] have contributed to the algorithm's historical injustice. In conjunction with developments regarding COMPAS's predictive methodology, legal and institutional changes must happen for COMPAS's just deployment.

## II. Introduction

Fairness is hard to measure, and there are many different ways to measure it. When training machine learning models, we are often interested in the perceived 'fairness' of those models, especially if they are being deployed to areas in which bias is especially sensitive. For instance, AI in judicial systems is heavily scrutinized for consistent rulings or predictions, irrespective of race, gender, or sexual identity, as these are protected attributes.

Steps have been taken to design machine learning systems from the ground up that are able to measure, and self-adjust, to prioritize fairness, rather than simply accuracy. This in itself is not trivial; first, a definition of fairness must be agreed upon because models are unable to satisfy multiple definitions of fairness, as demonstrated by Arrow's impossibility theorem [7][9][11][12]. Then, we must determine to what extent the quality of the accuracy should suffer in pursuit of that fairness. Finally, biases within an AI algorithm's training data make it extremely difficult to make an AI algorithm, which inherits the bias within its historical training data, fairer.

This paper focuses on augmenting the fairness of Correctional Offender Management Profiling for Alternative Sanctions (COMPAS) [13], a carceral AI risk assessment that has been employed in U.S. courts since the early 2000s to assess a defendant's likelihood of recidivism and assist judicial decision-making processes. COMPAS has faced significant criticism because of its bias towards African Americans, disproportionately classifying them as high-risk [4][9][11]. This bias within COMPAS has caused a disproportionate amount of African Americans to endure harsher sentences, parole denial, worsened future job prospects, the loss of voting rights, limited educational opportunities, and increased difficulty in securing housing [4][9]. There are two primary causes of COMPAS's bias: societal systematic bias and algorithmic logic. Historical societal biases cause bias within AI algorithms' training data—highlighted by the historical cycle of over-policing and arresting of African Americans. Also, the algorithmic logic of COMPAS affects its fairness because its specific optimization of fairness and accuracy impacts its classifications. This paper concentrates on improving COMPAS's algorithmic logic, proposing a novel approach to comparing the fairness of AI algorithms' optimizations and improving the fairness of COMPAS with the GAP framework as proposed by Gupta et al. (2024) [1].

## III. Background

There has been a lot of previous work on defining the fairness of AI algorithms. Various traditional fairness metrics have been created and are commonly used in machine learning. Some examples are shown in Figure A.

| Fairness Notion | Label | Formulation |
|---|---|---|
| Equalized Odds | f1 | $\frac{1}{2} * (|\frac{FP_0}{FP_0+TN_0} - \frac{FP_1}{FP_1+TN_1}| + |\frac{TP_0}{TP_0+FN_0} - \frac{TP_1}{TP_1+FN_1}|)$ |
| Error difference | f2 | $\frac{FP_0+FN_0}{N_1+N_0} - \frac{FP_1+FN_1}{N_1+N_0}$ |
| Error ratio | f3 | $\frac{\frac{FP_0+FN_0}{N_1+N_0}}{\frac{FP_1+FN_0}{N_1+N_0}}$ |
| Discovery difference | f4 | $\frac{FP_0}{TP_0+FP_0} - \frac{FP_1}{TP_1+FP_1}$ |
| Discovery ratio | f5 | $\frac{\frac{FP_0}{TP_0+FP_0}}{\frac{FP_1}{TP_1+FP_1}}$ |
| Predictive Equality | f6 | $\frac{FP_0}{FP_0+TN_0} - \frac{FP_1}{FP_1+TN_1}$ |
| FPR ratio | f7 | $\frac{\frac{FP_0}{FP_0+TN_0}}{\frac{FP_1}{FP_1+TN_1}}$ |
| False Omission rate (FOR) difference | f8 | $\frac{FN_0}{TN_0+FN_0} - \frac{FN_1}{TN_1+FN_1}$ |
| False Omission rate (FOR) ratio | f9 | $\frac{\frac{FN_0}{TN_0+FN_0}}{\frac{FN_1}{TN_1+FN_1}}$ |
| Disparate Impact | f10 | $\frac{\frac{TP_0+FP_0}{N_0}}{\frac{TP_1+FP_1}{N_1}}$ |
| Statistical Parity | f11 | $\frac{TP_0+FP_0}{N_0} - \frac{TP_1+FP_1}{N_1}$ |
| Equal Opportunity | f12 | $\frac{TP_0}{TP_0+FN_0} - \frac{TP_1}{TP_1+FN_1}$ |
| FNR difference | f13 | $\frac{FN_0}{FN_0+TP_0} - \frac{FN_1}{FN_1+TP_1}$ |
| FNR ratio | f14 | $\frac{\frac{FN_0}{FN_0+TP_0}}{\frac{FN_1}{FN_1+TP_1}}$ |
| Average odd difference | f15 | $\frac{1}{2} * (\frac{FP_0}{FP_0+TN_0} - \frac{FP_1}{FP_1+TN_1} + \frac{TP_0}{TP_0+FN_0} - \frac{TP_1}{TP_1+FN_1})$ |
| Predictive Parity | f16 | $\frac{TP_0}{TP_0+FP_0} - \frac{TP_1}{TP_1+FP_1}$ |

Figure A [7]

FP represents a false positive, which occurs when the model incorrectly predicts the outcome to be positive. By similar logic, FN represents a false negative, which occurs when the model incorrectly predicts the outcome to be negative. TP represents a true positive, which occurs when the model correctly predicts the outcome to be positive, and TN represents a true negative, which occurs when the model correctly predicts the outcome to be negative. N represents the sample size of each group.

The subscripts of 0 and 1 distinguish different groups' metrics. In the case of this study, the two groups that this paper focuses on are African Americans and Caucasians. For the majority of the metrics in the table, by definition, optimal fairness is achieved when the difference is 0 or when the ratio is 1. While optimizing for perfect fairness is possible, fairness and accuracy are often indirectly proportional, causing the model's accuracy to be greatly harmed and necessitating a balance between fairness and accuracy in most cases.

$$p(\hat{y} \neq y | y = 1, s) = p(\hat{y} \neq y | y = 1)$$

$$p(\hat{y} \neq y | y = 0, s) = p(\hat{y} \neq y | y = 0)$$

Defining the paper's statistical notation, y is the actual outcome, with y = 1 representing a positive outcome and y = 0 representing a negative outcome. ŷ is the predicted outcome, with ŷ = 1 representing a positive model prediction and ŷ = 0 representing a negative model prediction.

With the increasing prominence of AI in judicial settings throughout the past two decades, numerous researchers have attempted to apply these various fairness metrics to COMPAS, reconstructing COMPAS and optimizing the model to prioritize these definitions. In virtually all of the research studies completed with regard to this topic, COMPAS's predictive accuracy is harmed with the prioritization of fairness, in its various definitions, as a result of the inverse relationship between fairness and accuracy. After this experimentation process, most researchers have concluded that the new balance between fairness and accuracy that they have produced is not clearly more desirable than COMPAS's current implementation.

This paper explores a novel fairness framework known as Group Accuracy Parity (GAP) [1]. GAP is a differentiable loss function that measures group accuracy with Cross-Entropy (CE), a differentiable loss function that measures the varying predictive accuracies of certain groups and minimizes the gap between this accuracy gap by minimizing Accuracy Difference (AD) to achieve Accuracy Parity (AP), a fairness definition that idealizes balanced detection accuracy across groups. The minimization of CE results in the minimization of Overall Error (OE). Binary Cross Entropy (BCE) is a loss function used for binary classifier optimization and serves as the foundation of GAP. Since BCE fails to take into account COMPAS's label imbalance, this paper's implementation of GAP utilizes Weighted Binary Cross Entropy (wBCE) as its Cross-Entropy metric to account for these class imbalances.

$$AD = \underbrace{P[\hat{y} = y | g = 1]}_{\text{Acc Group 1 (g=1)}} - \underbrace{P[\hat{y} = y | g = 0]}_{\text{Acc Group 0 (g=0)}}$$

$$BCE = -\frac{1}{N} \sum_{N} y \log(\hat{y}) + (1 - y) \log(1 - \hat{y})$$

$$wBCE = -\frac{1}{N} \sum_{N} w(y) \cdot y \log(\hat{y}) + w(1 - y) \cdot (1 - y) \log(1 - \hat{y})$$

$$GAP = OE + \lambda \sum_{i,j \in [G], i \neq j} \| \underbrace{CE(g = i)}_{\text{err Group i (g=i)}} - \underbrace{CE(g = j)}_{\text{err Group j (g=j)}} \|_2^2$$

Figure B [1]

**IV. Dataset**

  The dataset that this paper explores is Correctional Offender Management Profiling for Alternative Sanctions (COMPAS) [13], a carceral AI risk assessment that has been employed in U.S. courts since the early 2000s to assess a defendant's likelihood of recidivism and assist judicial decision-making processes.
  This paper explores a database, released by ProPublica [2], of defendants from Broward County, Florida that have been affected by COMPAS. The database includes many statistics, such as defendants' criminal history, race, sex, the risk scores that COMPAS predicted, and whether or not they actually recidivated two years after their release.
  The dataset includes 7214 defendants, primarily consisting of African American and Caucasian defendants, which are the 2 races that have enough data points for us to analyze. The other racial groups are too small.

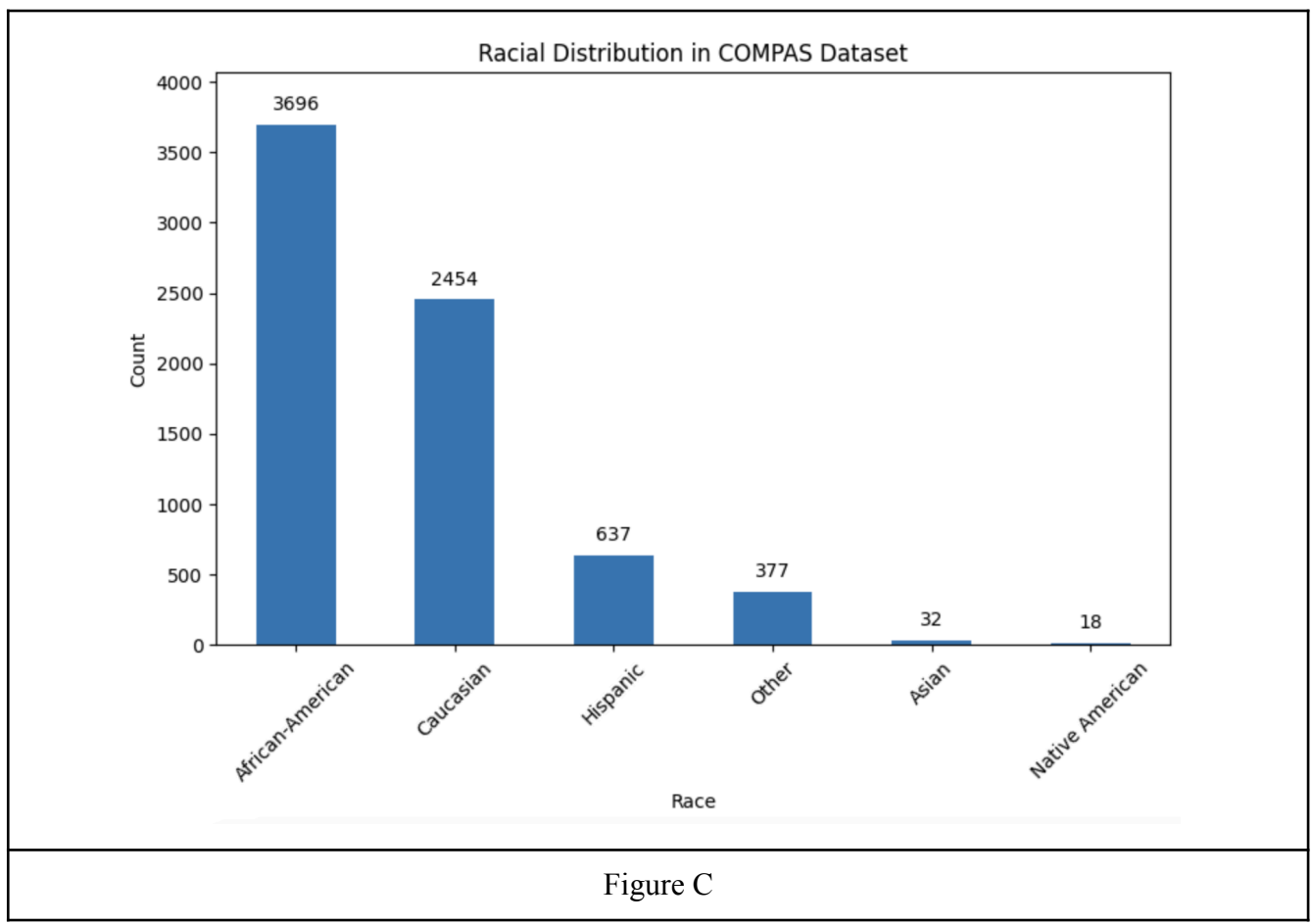

Figure C

This paper also considers the protected attribute of sex in its analysis because both sex groups have enough data points.

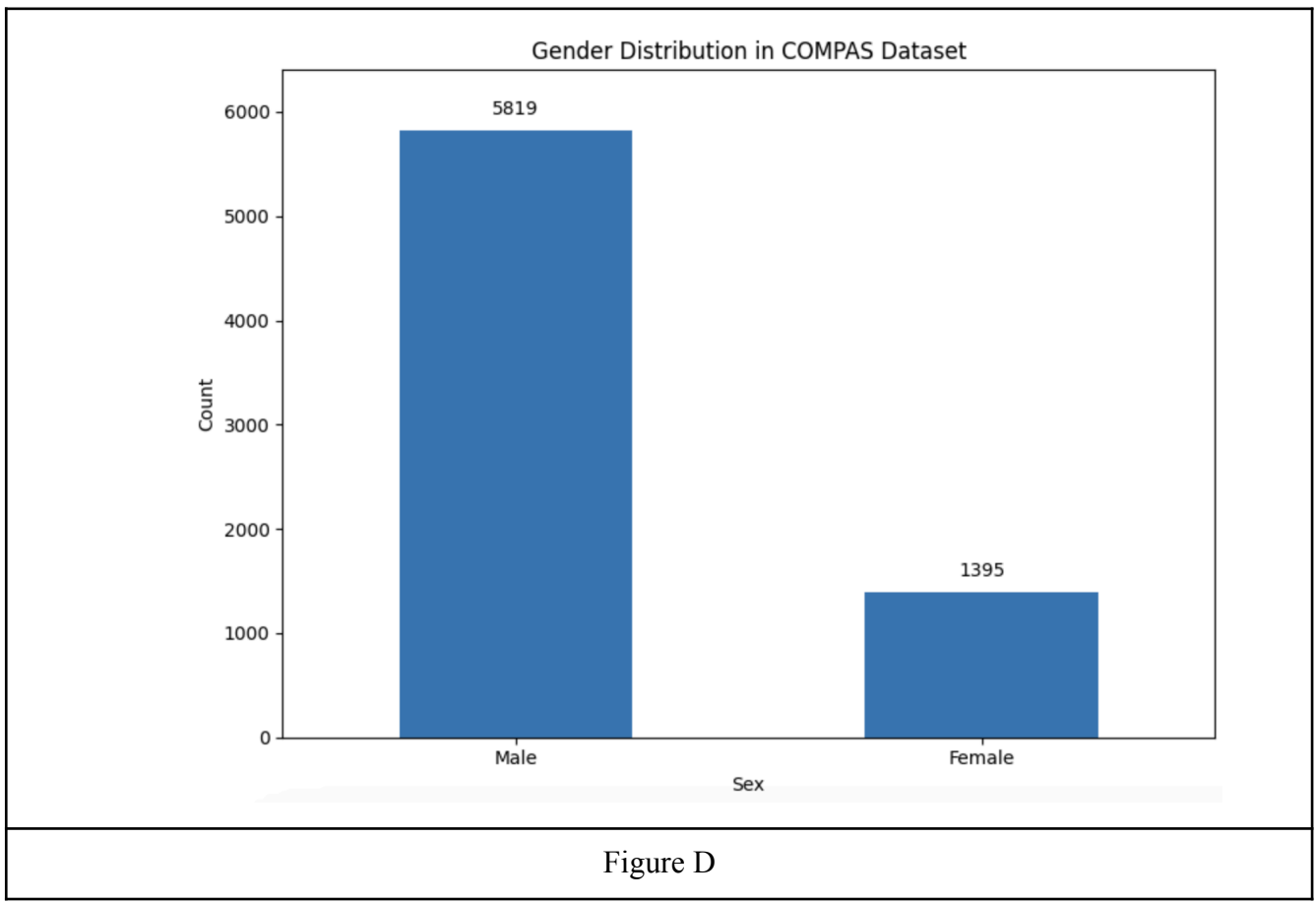

Figure D

As for the age groups of defendants in the dataset, this paper focuses on defendants between the ages of 18 and 40 because this range is where most of the data points are.

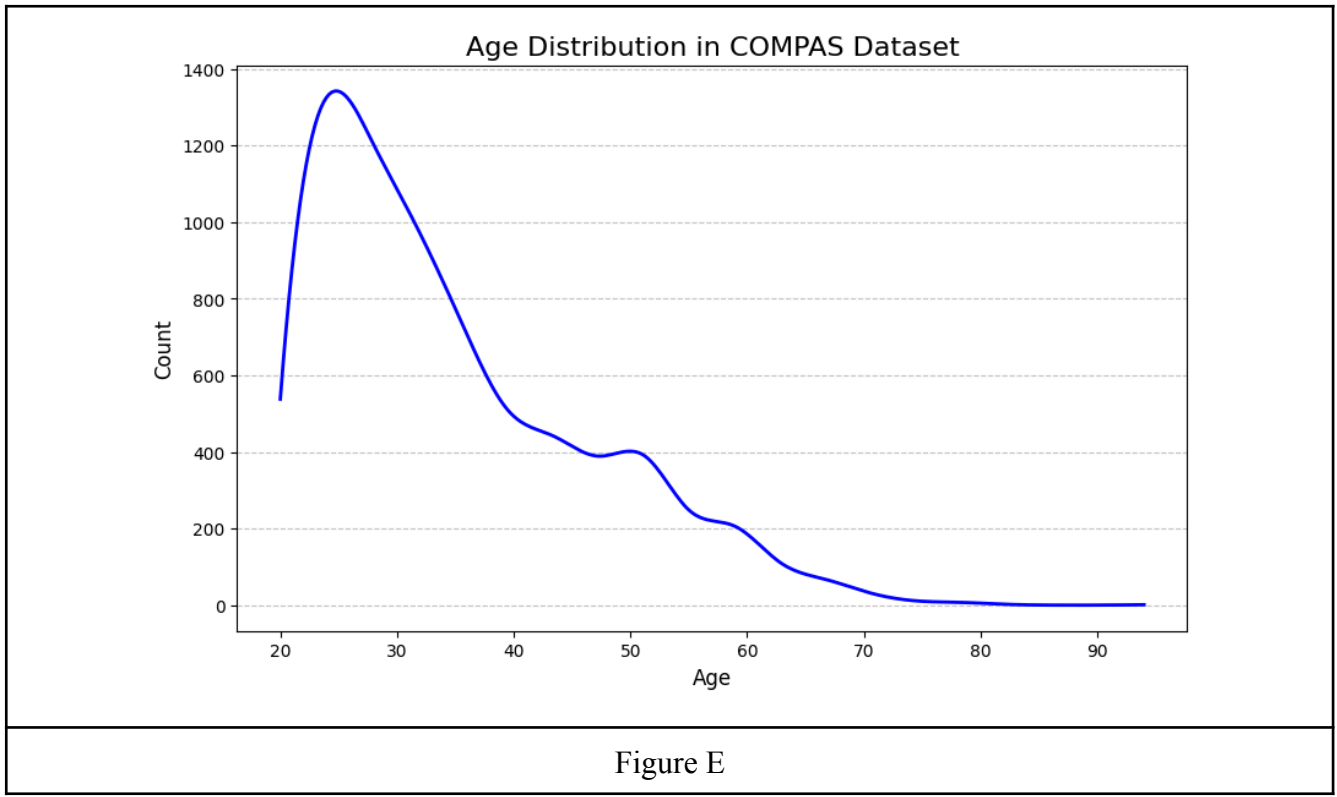

Figure E

**V. Methodology/Models**

      In order to observe the effects of using a custom loss function that prioritizes fairness, and to help define a systematic methodology through which fairness can be investigated, the COMPAS dataset was used as a benchmark.

      In order to properly differentiate the results of custom functions against generic ones, a full suite of machine learning models was used to give a generic baseline, from which the custom methodology aims to improve upon. The COMPAS dataset was chosen as it is famous for the fact that while the accuracy between the two sensitive attributes of African American and Caucasian were similar, the relative false positive rates between the groups skewed heavily towards African Americans. In essence, although the model was accurate at the same rates between races, the form of error was not consistent, and the outcome for African American defendants was systematically worse than their Caucasian peers. This is exactly the kind of environment in which this paper proposes that a GAP-powered loss function would be deployed, as it allows for the useful features of the model, while explicitly training for equal outcomes between sensitive groups.

      The introduction of such a loss function was weighed between existing loss functions across different types of models. Selection of any hyperparameter, such as a loss function, is both non-trivial and varies depending on the nature of the data and its use case. For instance, within a linear regression, a 'mean squared error' may be preferred in some cases over a 'mean absolute error' depending on the underlying data and the kind of trend that is most appropriate. Both are

sound, scientifically relevant functions, and opinions on which is correct may well vary from scientist to scientist.

Model selection for these tasks is also important, but difficult decisions lie in the tradeoff between interpretability and accuracy. A highly interpretable model may be easily explained and examined for fairness within its methodology, but could result in significantly less accurate results, which in itself could be viewed as 'unfair.' Conversely, a highly accurate model may be completely uninterpretable and, despite the high accuracy, questions about fairness remain because it is unclear how the model derived its conclusions and could be making decisions in a way that we would deem to be inappropriate.

Complex tasks increasingly rely on neural networks and other 'black box' models that have been highly scrutinized due to the fact that it is difficult to verify how any given decision was made. To this end, recent literature has focused on the ability of models to be trained from the ground up such that their definition of success is based on what is fair; or at least one of the many mathematical definitions of fairness. This has presented mathematical challenges as loss functions for neural networks must be differentiable - that is, must produce a continuous function such that small changes within the network can produce infinitely small increases or decreases in model score, such that fine-tuning of the network can take place. Traditional fairness metrics, as discussed in the Background section, often rely on some ratio of, for example, False Positives against False Negatives across sensitive categories. These ratios are not discrete as they are fundamentally bound by the total sample size of the training data, which are integers. Therefore, a loss function that uses such mathematical definitions of fairness directly would not be usable within a neural network. Nor would they be any good - a model that was solely focused on fairness may conclude that an optimal strategy would be to have no False Positives by assigning all testing data a negative label, thereby creating a 'fair' yet highly inaccurate model.

The Gupta et al. paper modified a traditional neural network loss function, cross-entropy (CE) to include a group accuracy parity component. Using Python, this paper implements GAP by creating a neural network trained on the COMPAS dataset. The original implementation used the GAP function to ensure that the accuracy between groups being targeted for hate speech was identified at the same rate. For instance, it was roughly as likely to identify hate speech targeting a Pacific Islander as it was for Native American. This model expands upon that research by analyzing the effect of the GAP in the full light of existing fairness functions that examine the differences in rate between False Positives and False Negatives. Existing models without GAP already achieved similar accuracies between the African American and Caucasian subsets, but the types of error that it made for those groups were not equal, and therefore the outcomes of the different groups were not fair. This GAP approach to COMPAS is the first of its kind and expands from Gupta et al's paper by primarily examining the quality of result not just by Group Accuracy Parity, but by Group Fairness Parity.

## VI. Results and Discussion

After its successful implementation and application of the GAP framework with COMPAS, this paper realizes that analyzing exactly how well it works is non-trivial.

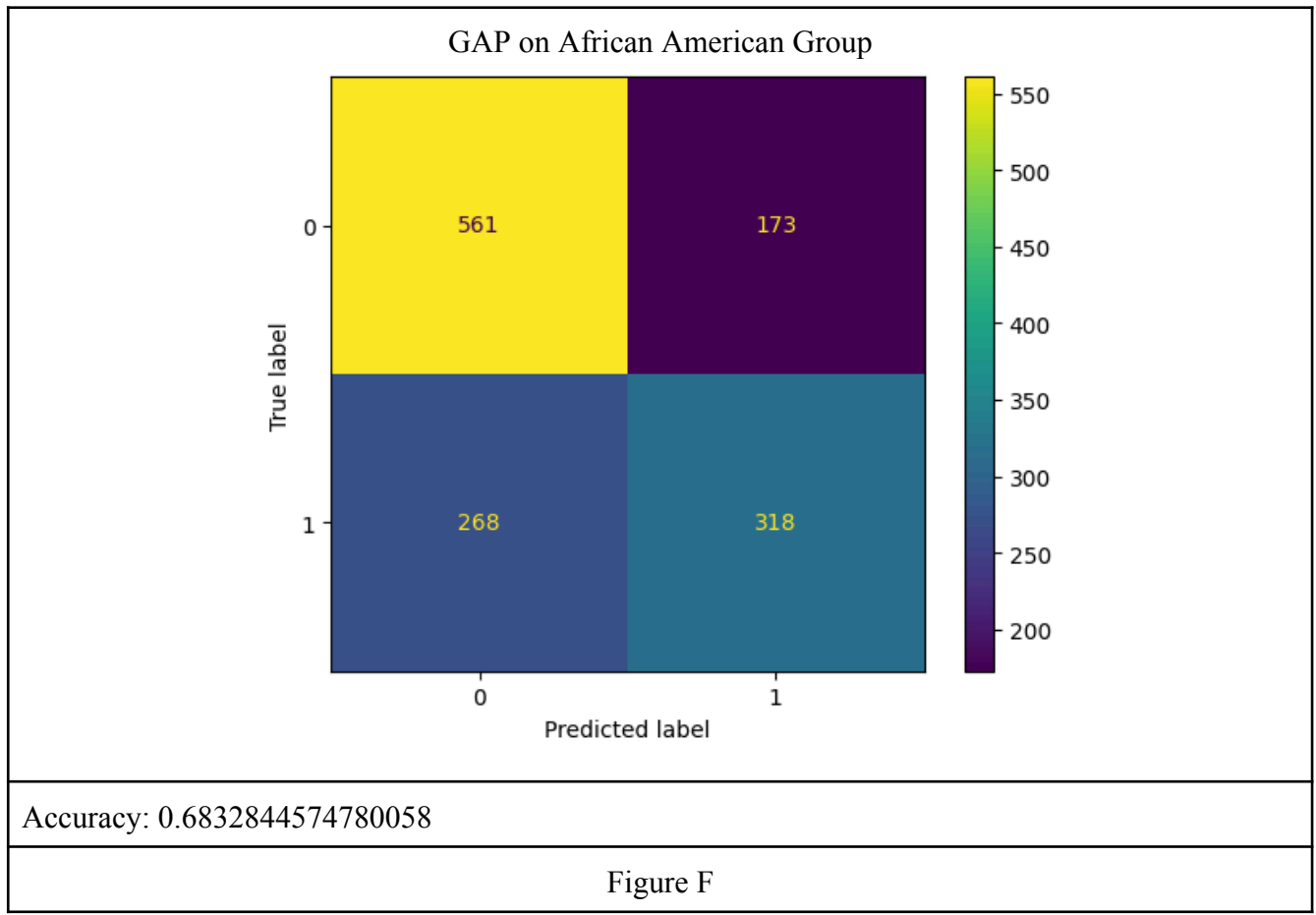

Accuracy: 0.6832844574780058

Figure F

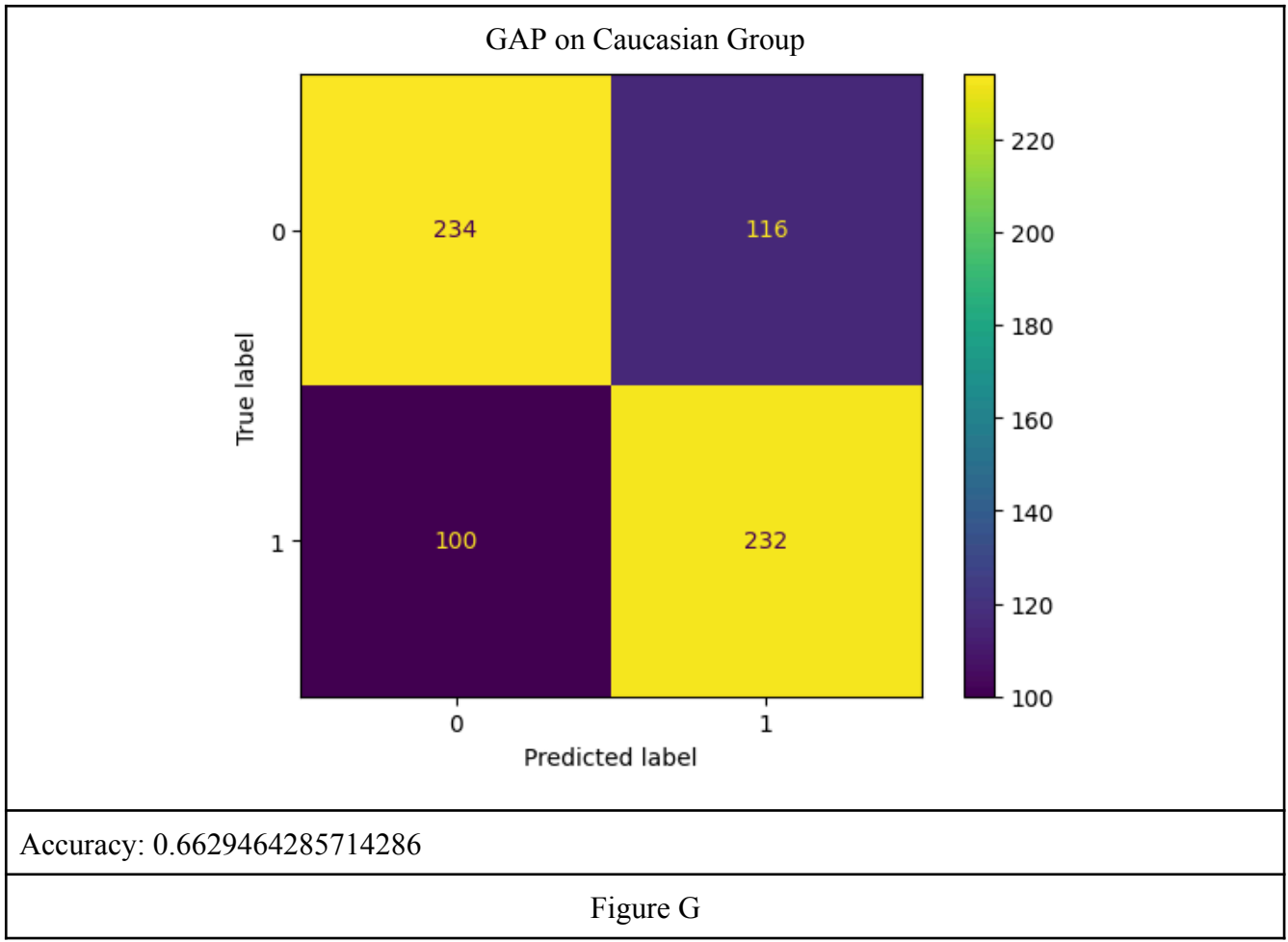

Accuracy: 0.6629464285714286

Figure G

    From these confusion matrices, GAP appears to be fairer based on the original complaints of COMPAS; however, as evidenced by a comparison of GAP's confusion matrices to an Adam optimizer on binary cross entropy confusion matrix, defining exactly how fair a model truly is certainly requires more than superficial confusion matrices. On an important note, the paper discovers that GAP is highly volatile, requiring numerous iterations of trials to achieve the confusion matrices shown in Figure F and Figure G. This observation indicates that we may need more data.

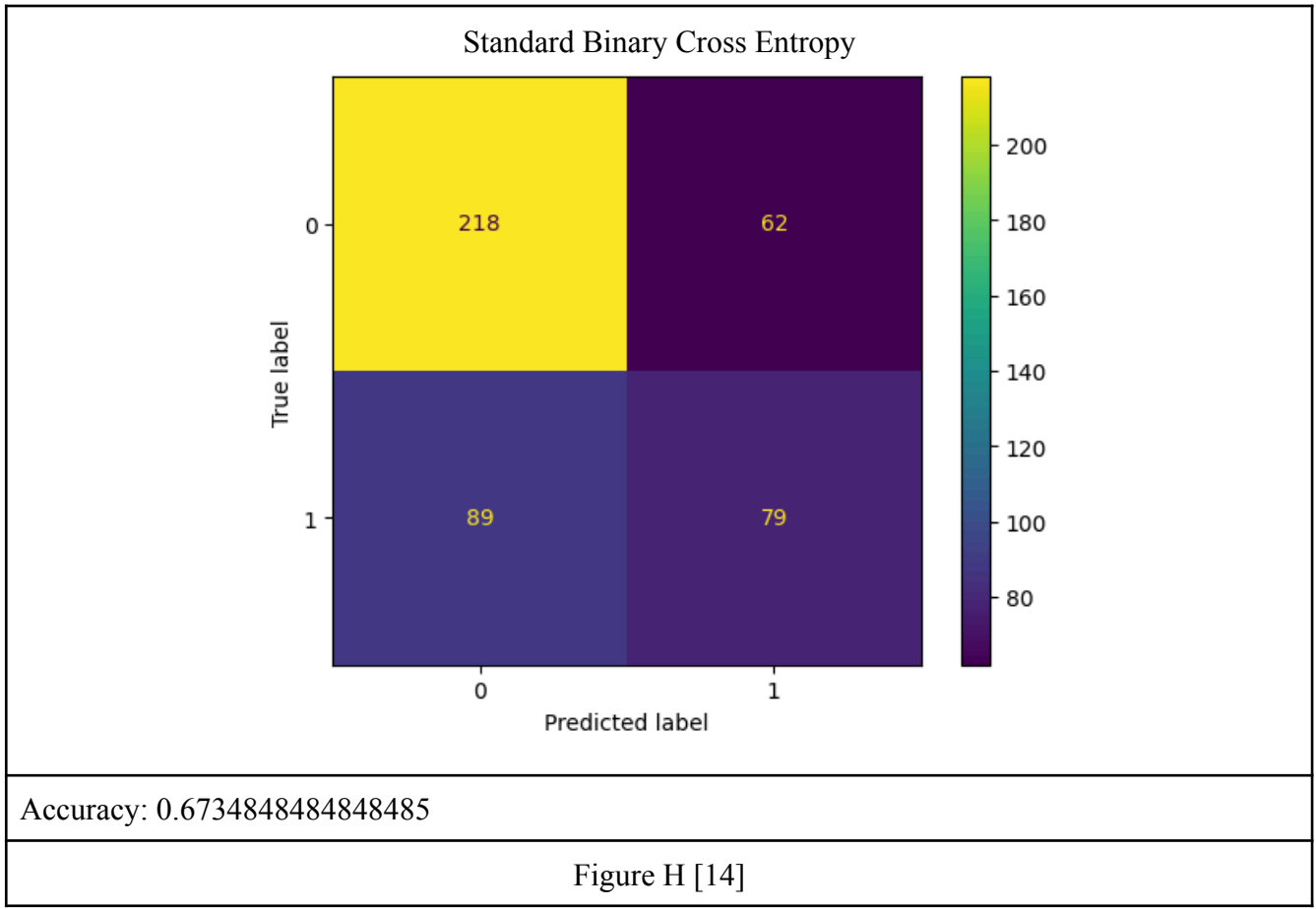

| |
|---|
| Accuracy: 0.6734848484848485 |
| Figure H [14] |

Hence, this paper proposes a novel procedure to investigate the complicated nature of this problem and its corresponding variables.

The definition of fairness depends not just on the problem but on the dataset. The dataset and the question being asked will determine which variables are more appropriate to use and whether a given loss function properly accounts for the existing trends and correlations within the data.

Additionally, the question of variable independence arises. In its analysis of COMPAS, this paper determines that there are many proxies for race.

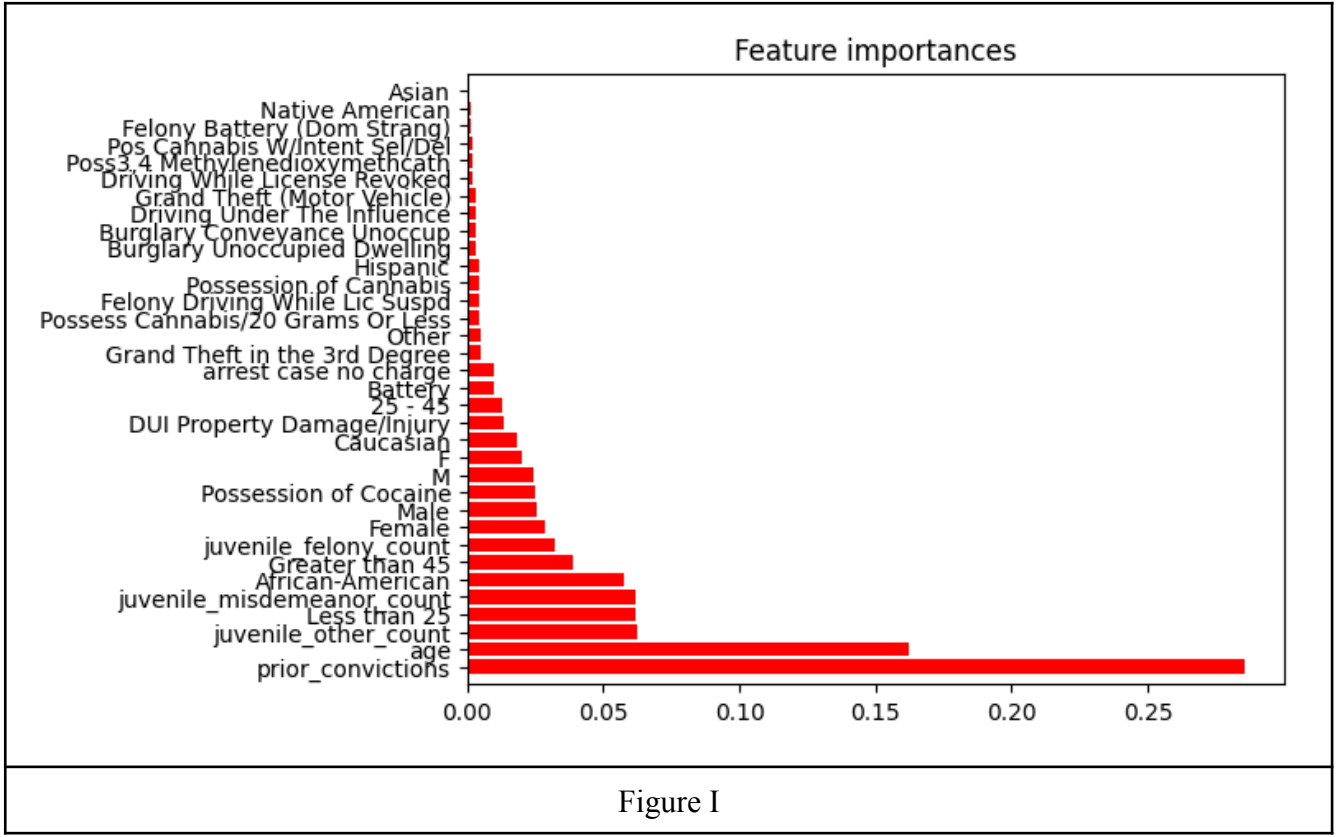

Figure I

The paper's proposed methodological methodology begins with a Pareto frontier analysis. We use the baseline of a Pareto frontier visualization to effectively quantify and compare different fairness metrics. This Pareto frontier baseline accuracy is the predictive accuracy of the model when it is 100% fair, according to what optimal fairness means in that specific fairness definition. Thus, the accuracy-fairness tradeoff of each fairness framework, defined by its own mathematical optimizations, is quantifiable by evaluating the baseline of the Pareto frontier. For demonstration, Figure J shows a Pareto frontier of a COMPAS model optimized to prioritize equalized odds (EOd), wherein the accuracy of the model at perfect fairness is roughly 0.52.

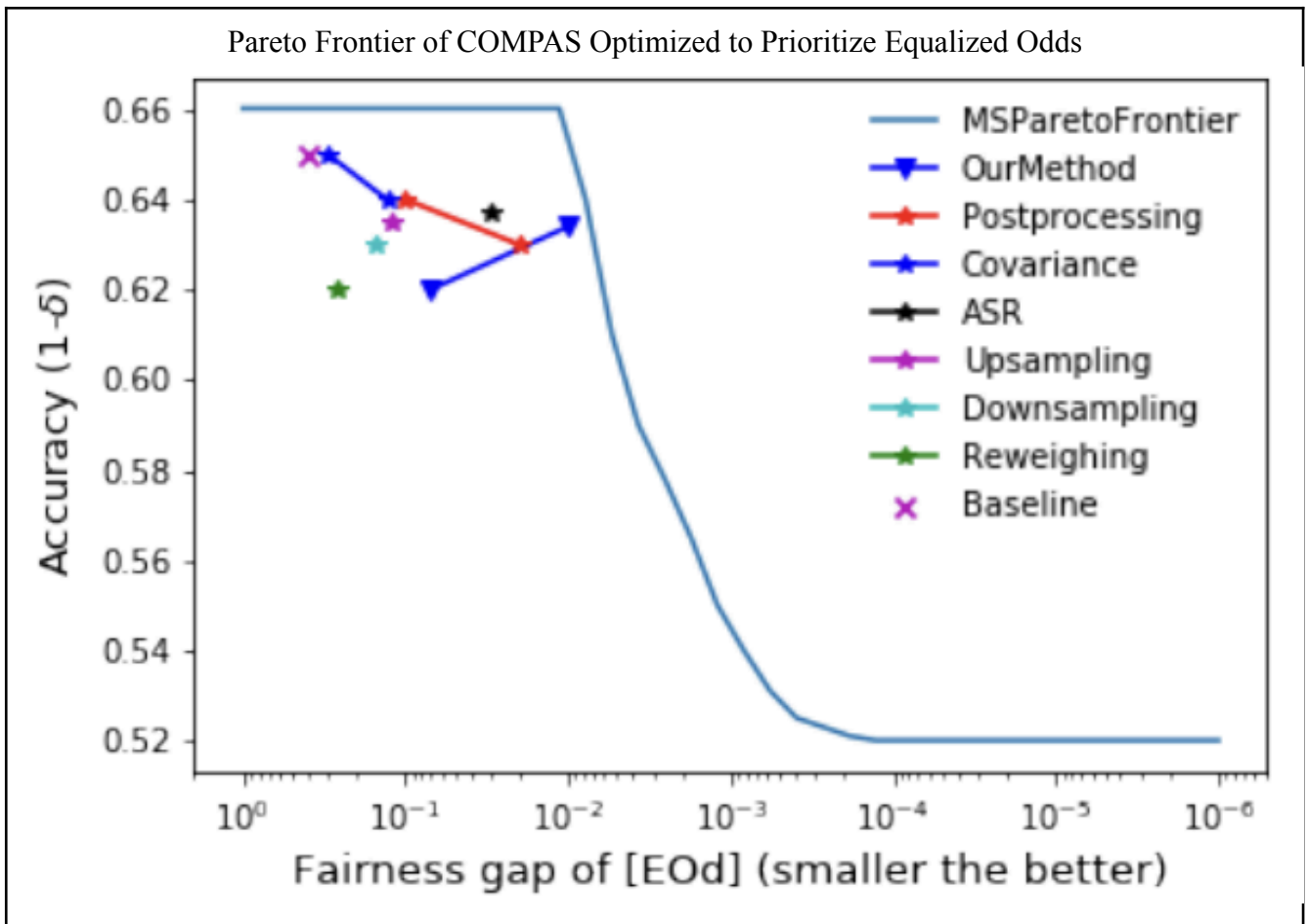

Figure J [3][15]

While the Pareto frontier is a good metric to analyze complex accuracy-fairness tradeoffs, it is not entirely reliable because different variables are often proxies of each other. If the dataset consists of considerable multivariable correlation, the Pareto frontier becomes less reliable because the fairness metric measures disparities introduced by this variable interdependence. The tradeoffs within the Pareto frontier, as a result, represent tradeoffs that are not truly independent from each other.

Therefore, the next step in the paper's proposed methodology allows us to investigate the complex relationships between the dataset's variables, the dataset's protected variables, and the loss function of choice. The methodology utilizes violin graphs to visualize the relationships between variables.

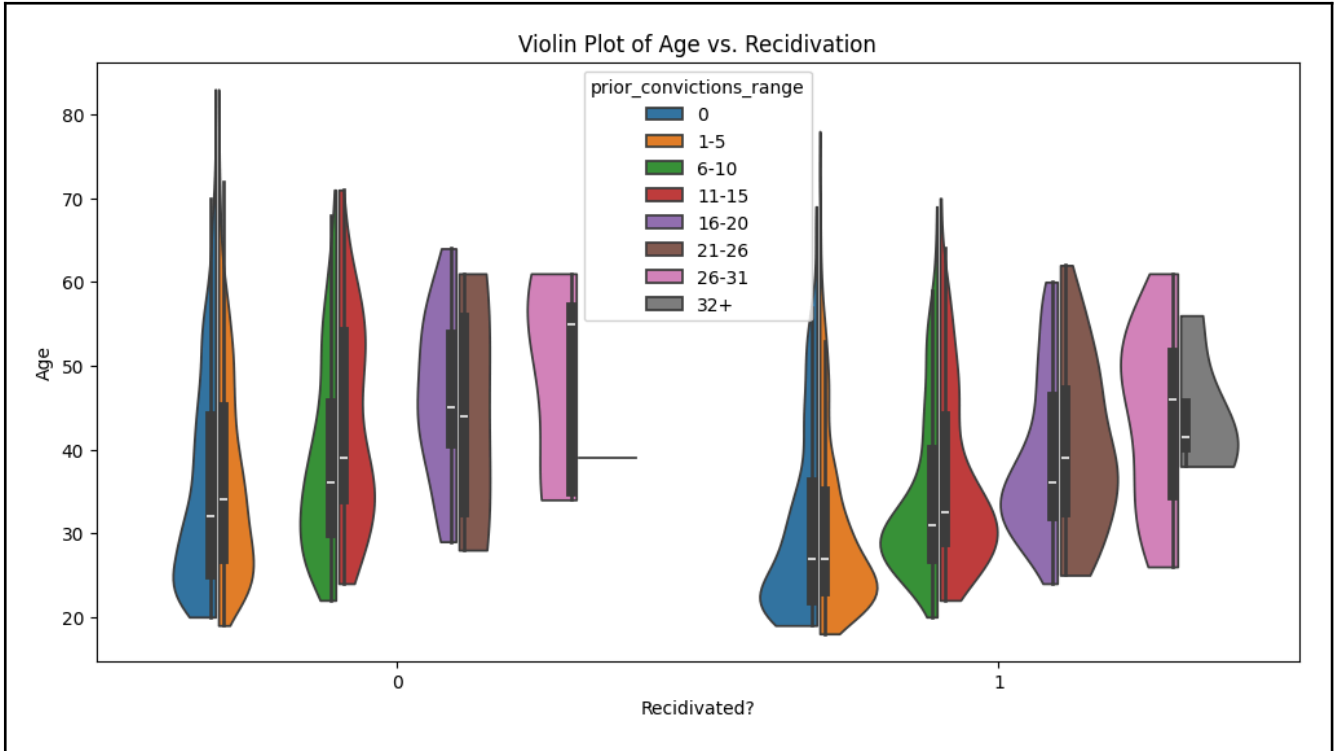

Figure K

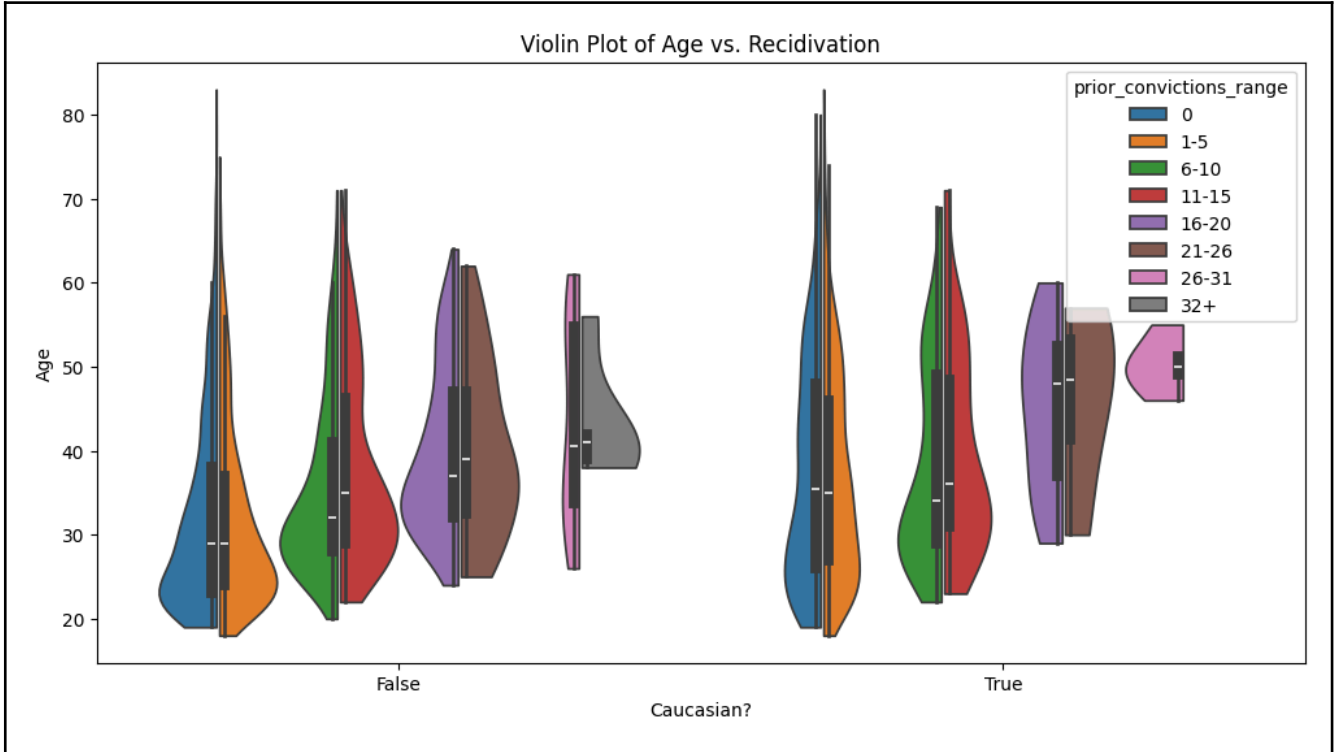

Figure L

The similarity between violin graphs fundamentally tells us how much of a proxy one variable is for another. The violin graph of the African American defendants is extremely similar to the violin graph of the recidivated group. Furthermore, the violin graph of the Caucasian defendants and the violin graph of the non-recidivated group. Under a court of law, age and prior convictions, which are the two factors that the first violin plot explores, would be deemed fair. However, the patterns present in the Figure K violin plot, while it does not include any race-specific data, evidently serve as an almost-perfect proxy for race. Clearly, simply removing race from COMPAS does not eliminate its racial bias. COMPAS's efficiency in predicting whether or not a defendant recidivates in the last two years is extremely similar to its efficiency in being African American or Caucasian.

The paper argues that the optimization of a fairness framework like GAP is necessary because the attunement of accuracy is unimportant relative to the critical problem of different variables serving as proxies for protected attributes like race.

On an important note, this paper only tests GAP. While the paper is unsure of whether GAP is the best choice of a fairness framework, its implementation of GAP shows that the fairness framework is implementable. With varying datasets, different mathematical frameworks make more sense for different instances.

Therefore, in the future, we aim to create a library containing loss functions for neural networks trained on different loss functions. This would be a system by which we can design fairer AI models.

However, building an AI model that is fairer than a human, which was the primary goal of COMPAS's implementation in judicial settings, inevitably requires some level of human decision because deciding which variables for the algorithm to evaluate and designing its internal methodology necessitates some human input. It is impossible to completely mathematically abstract the implementation of algorithms as computer scientists, raising the question of whether we can ever beat judges. The model requires an assessment of the optimal fairness, which is the responsibility of the judicial judges. It is difficult to make a model fairer than a human when the creation of the model requires human input.

Perhaps, we can create a voting system for people to vote on which framework instead of a single judge making all the decisions. A comprehensive system that details all of the fairness frameworks and metrics would greatly facilitate this process.

## VII. Conclusion

This paper systematically examines fairness parity by using a custom loss function relying on the fairness properties of GAP. Through this application of GAP, this paper has shown that the framework results in an increase in COMPAS's fairness. This paper highlights the fundamental flaws in approaching absolute fairness metrics for absolute datasets, particularly in complex, real-world scenarios where variables are often proxies of each other. Considerable variable interdependence complicates the process of analyzing accuracy-fairness tradeoffs because it prevents an increase in accuracy without effectively discovering some pattern that is present between both non-protected and protected variables. Consequently, even with a custom-loss function, the derivation of accuracy from fairness is typically problematic because what makes a model accurate might concurrently make it unfair. While this relationship is hugely important and has powerful, real-world consequences, there has yet to be much discussion regarding this topic in literature.

## VIII. Acknowledgements

I would like to thank Simeon Sayer from Harvard University for his expert mentorship on this research project.

## IX. References

**[1]** Gupta, S., Kovatchev, V., De-Arteaga, M., & Lease, M. (2024). Fairly accurate: Optimizing accuracy parity in fair target-group detection. *arXiv preprint arXiv:2407.11933*.

**[2]** Mattu, J. L. a. K. (2023, December 20). How we analyzed the COMPAS Recidivism Algorithm. ProPublica. https://www.propublica.org/article/how-we-analyzed-the-compas-recidivism-algorithm

**[3]** Chai, J., & Wang, X. (2022, June). Fairness with adaptive weights. In International Conference on Machine Learning (pp. 2853-2866). PMLR.

**[4]** Rudin, C., Wang, C., & Coker, B. (2020). The age of secrecy and unfairness in recidivism prediction. Harvard Data Science Review, 2(1), 1.

**[5]** Hall, P., & Gill, N. (2017). Debugging the black-box COMPAS risk assessment instrument to diagnose and remediate bias.

**[6]** Alikhademi, K., Richardson, B., Drobina, E., & Gilbert, J. E. (2021). Can explainable AI explain unfairness? A framework for evaluating explainable AI. arXiv preprint arXiv:2106.07483.

**[7]** Anahideh, H., Nezami, N., & Asudeh, A. (2021). On the choice of fairness: Finding representative fairness metrics for a given context. arXiv preprint arXiv:2109.05697, 1-25.

**[8]** Heaven, W. D. (2020). Predictive policing algorithms are racist. They need to be dismantled. MIT Technology Review, 17, 2020.

[9] Thomas, S. (2023). The Fairness Fallacy: Northpointe and the COMPAS Recidivism Prediction Algorithm (Doctoral dissertation, Columbia University).

[10] Washington, A. L. (2018). How to argue with an algorithm: Lessons from the COMPAS-ProPublica debate. Colo. Tech. LJ, 17, 131.

[11] Humerick, J. D. (2019). Reprogramming fairness: Affirmative action in algorithmic criminal sentencing. HRLR Online, 4, 213.

[12] Ruf, B., & Detyniecki, M. (2021). Towards the right kind of fairness in AI. arXiv preprint arXiv:2102.08453.

[13] Mattu, J. a. L. K. (2023, December 20). Machine bias. ProPublica. https://www.propublica.org/article/machine-bias-risk-assessments-in-criminal-sentencing

[14] Jain, B., Huber, M., & Elmasri, R. (2021). Increasing fairness in predictions using bias parity score based loss function regularization. arXiv preprint arXiv:2111.03638.

[15] Kim, J. S., Chen, J., & Talwalkar, A. (2020, November). FACT: A diagnostic for group fairness trade-offs. In International Conference on Machine Learning (pp. 5264-5274). PMLR.